\pdfoutput=1

\documentclass[11pt]{article}

\usepackage{EMNLP2023}

\usepackage{times}
\usepackage{latexsym}

\usepackage[T1]{fontenc}

\usepackage[utf8]{inputenc}

\usepackage{microtype}

\usepackage{inconsolata}
\usepackage{microtype}
\usepackage{amsmath}
\usepackage{amssymb}
\usepackage{mathtools}
\usepackage{amsthm}
\DeclareMathOperator*{\argmin}{argmin}
\usepackage{hyperref}
\usepackage{multirow}
\usepackage{tabularray}
\usepackage{booktabs}
\usepackage{times}
\usepackage{latexsym}
\usepackage{comment}

\title{An Empirical Bayes Framework for Open-Domain Dialogue Generation}

\author{Jing Yang Lee\textsuperscript{1}, Kong Aik Lee\textsuperscript{2}, Woon Seng Gan\textsuperscript{3}\\\\
  School of Electrical and Electronic Engineering, Nanyang Technological University\textsuperscript{1,3} \\
  Institute for Infocomm Research, A*STAR\textsuperscript{2}\\
  jingyang001@e.ntu.edu.sg\textsuperscript{1}, lee\_kong\_aik@i2r.a-star.edu.sg\textsuperscript{2}, ewsgan@ntu.edu.sg\textsuperscript{3} \\}

\begin{document}
\maketitle
\begin{abstract}
To engage human users in meaningful conversation, open-domain dialogue agents are required to generate diverse and contextually coherent dialogue. Despite recent advancements, which can be attributed to the usage of pretrained language models, the generation of diverse and coherent dialogue remains an open research problem. A popular approach to address this issue involves the adaptation of variational frameworks. However, while these approaches successfully improve diversity, they tend to compromise on contextual coherence. Hence, we propose the Bayesian Open-domain Dialogue with Empirical Bayes (BODEB) framework, an empirical bayes framework for constructing an Bayesian open-domain dialogue agent by leveraging pretrained parameters to inform the prior and posterior parameter distributions. Empirical results show that BODEB achieves better results in terms of both diversity and coherence compared to variational frameworks.
\end{abstract}

\section{Introduction}
An open-domain dialogue agent, or chatbot, aims to engage users in meaningful conversation by generating diverse, contextually coherent dialogue. In recent years, advances in deep learning and parallel computing have resulted in significant progress in open-domain dialogue research. Architectures incorporating large transformer-based Pretrained Language Models (PLMs) such as BERT \cite{devlin-etal-2019-bert}, T5 \cite{https://doi.org/10.48550/arxiv.1910.10683}, and GPT-3 \cite{https://doi.org/10.48550/arxiv.2005.14165} have achieved state of the art performance. However, despite recent progress, the fundamental issues of response diversity persists. Open-domain dialogue agents still demonstrate a tendency to generate repetitive, generic responses. 

To address this issue, a broad range of approaches have been proposed. In recent years, however, the usage of variational, or latent variable, frameworks has emerged as the most popular approach. Variational approaches broadly involve the application of variational auto-encoding frameworks such as the Variational Auto Encoder (VAE) \cite{https://doi.org/10.48550/arxiv.1312.6114}, Wasserstein Auto Encoder(WAE) \cite{https://doi.org/10.48550/arxiv.1711.01558}, and the Conditional Variational Auto Encoder (CVAE) \cite{NIPS2015_8d55a249} in particular, to open-domain dialogue. Essentially, these approaches involve modelling the one-to-many relationship of dialogue by defining and randomly sampling a latent multivariate Gaussian prior distribution during response generation \cite{ijcai2022p564}. However, while the stochasticity induced via randomly sampling the latent prior successfully improves response diversity, the coherence of the responses often suffer as the sampled latent variables fail to accurately capture the semantics of the dialogue context \cite{lee-etal-2022-randomized}.

In order to improve diversity while maintaining coherence, we turn to Bayesian deep learning. Traditionally, deep learning involves performing inference and optimization on a real-valued, deterministic neural network. For Bayesian deep learning, learning and inference are performed on a Bayesian Neural Network (BNN), which regards each model parameter as a Gaussian distribution \cite{journals/neco/MacKay92a}. Essentially, a prior distribution is specified for each weight or bias, and Bayesian inference is performed to attain the posterior distribution. During inference, each Gaussian posterior is randomly sampled to attain a single weight or bias value. In other words, the BNN can be also be viewed as an ensemble of models. Hence, instead of relying on randomly sampling a single latent prior distribution, stochasticity is introduced when each bayesian parameter is sampled. Similar to variational frameworks, a BNN would also naturally model the one-to-many property of dialogue.

However, training a BNN from scratch for open-domain dialogue would be challenging. For BNNs, the choice of prior is vital to model performance. Selecting an uninformative, vague prior would result in poor performance. In the context of open-domain dialogue, it would be exceedingly difficult to select a prior which accounts for the complexities of dialogue data. Moreover, achieving proficient natural language generation usually necessitates a relatively large model trained on a substantial amount of textual data. Additionally, the model size would be effectively doubled as each network parameter is now represented by a Gaussian defined by a mean and variance. This results in high computational cost and long training time.

Since PLMs have demonstrated strong language understanding and generation capabilities, we attempt to leverage their capabilities via an Empirical Bayes approach such as the MOdel Priors with Empirical Bayes using DNN (MOPED) framework \cite{Krishnan_Subedar_Tickoo_2020}. MOPED involves utilizing the deterministic parameters in a deterministic neural network to inform the mean of the prior, and both the mean and variance of the posterior in a BNN. However, directly applying MOPED to a PLM for open-domain dialogue generation presents challenges. This approach would not only double the already substantial number of parameters but also lead to a significant drop in contextual coherence. This drop occurs because of the excessive stochasticity introduced when each parameter is randomly sampled from the corresponding posterior distribution. Hence, we introduce the Bayesian Open-domain Dialogue with Empirical Bayes (BODEB) framework, inspired by MOPED but tailored specifically for open-domain dialogue generation. BODEB addresses the issue of excessive stochasticity, ensuring response coherence by:
\begin{enumerate}
    \item Only regarding \emph{selected} parameters in the model as Gaussian distributions.
    \item Incorporating information regarding the position of the parameter when defining the variance of both the prior and posterior.
    \item Utilising a mixture Gaussian prior with a spike-and-slab distribution instead of a standard Gaussian prior (a variant with a standard Gaussian prior is also presented).
\end{enumerate} 



To our knowledge, this is the first attempt at constructing a Bayesian open-domain dialogue agent where model parameters are modelled as probability distributions. We conduct extensive experiments on the DailyDialog \cite{li-etal-2017-dailydialog} and EmpatheticDialogs \cite{rashkin-etal-2019-towards} corpora. In our implementation, pretrained parameters from the GPT-2 \cite{radford2019language} and DialoGPT \cite{zhang-etal-2020-dialogpt} PLMs are used. Empirical results show that BODEB achieves better performance in terms of both diversity and coherence compared to variational frameworks. Additionally, we also conduct additional experiments to investigate the impact of Bayesian parameter selection and posterior variance on overall response diversity and contextual coherence.

\section{Background}

\noindent\textbf{Generative Open-Domain Dialogue} In this paper, we will focus on generative open-domain dialogue generation. Given a dialogue consisting of $K$ utterances, the input, also known as the dialogue context, $X$ consists of all prior utterances in the conversation (i.e., $X = \{ x_{1}, x_{2}, \cdots, x_{K-1} \}$). The label or reference response is simply the final utterance in the dialogue, $Y = x_{K}$. The agent, which features an encoder-decoder architecture, then generates the response $\bar{y}$ in an autoregressive manner.

In the context of variational open-domain dialogue agents \cite{zhao-etal-2017-learning, https://doi.org/10.48550/arxiv.2003.12738,9414586,9747458, shen-etal-2017-conditional, ijcai2020p503, https://doi.org/10.48550/arxiv.2207.12696, wu-etal-2020-guiding, zhou-wang-2018-mojitalk, 9406340}, during response generation, a latent variable $z$ is randomly sampled from a latent Gaussian prior distribution $p(z|X)$. The sampled latent variable $z$ is then fed to the decoder, which could consist of recurrent networks such as LSTM or GRUs, Transformer networks, or PLMs. During training, the latent variable is randomly sampled from an approximated posterior $p(z|X,\bar{y})$, where $\bar{y}$ represents the reference response. Both $p(z|X)$ and $p(z|X,\bar{y})$ are usually defined by an external networks. Parameters are optimized by minimizing the KL divergence between the latent prior $p(z|X)$ and the approximated posterior $p(z|X,y)$. This approach enhances response diversity through the stochastic nature of random sampling during response generation. Ideally, the latent variable $z$ captures the semantics related to potential dialogue response intents. However, due to the inherent complexity of open-domain dialogue, which exhibits both one-to-many and many-to-one phenomena \cite{sun-etal-2021-generating}, sampled latent variables often struggle to accurately capture contextual semantics. Consequently, this leads to a decrease in contextual coherence.

Alternative variational frameworks have also been designed specifically to address this issue. \citet{sun-etal-2021-generating} introduced the Self-separated CVAE which partitions the input data into a number of groups to reduce the disparity between dialogue contexts and latent variables. \citet{9747458} proposed the Uncertainty-Aware CVAE,a variant of the CVAE which incorporates an estimation of aleatoric uncertainty during inference. On the other hand, \citet{gao-etal-2019-discrete} and \citet{bao-etal-2020-plato} propose variational frameworks which utilize discrete latent variables instead of continuous latent variables. While these approaches do alleviate this issue to some extent, there is generally a compromise when it comes to diversity. 

\noindent\textbf{Bayesian Neural Networks} Bayesian Neural Networks (BNNs) provide a probabilistic interpretation of a standard neural network by representing each weight as a probability distributions over potential values. For this discussion $x$, $y$, and $\theta$ refer to the model inputs, outputs, and parameters (which consists of both weights and biases) respectively. A prior distribution over the network weights $p(\theta)$ is defined to capture any initial belief regarding which parameters would have likely generated the outputs. Subsequently, we aim to compute the posterior distribution using Bayes rule:
\begin{equation}
    p(\theta|x,y) = \frac{p(y|x,\theta)p(\theta)}{\int p(y|x,\theta)p(\theta)d\theta}
\end{equation}
where $p(y|x,\theta)$ is known as the likelihood, and the denominator represents the evidence. Due to the size of neural networks, computing the posterior $p(\theta|x,y)$ is usually intractable. In the context of BNNs, some popular approaches to attain a reliable approximation of the posterior include Hamiltonian Mone-Carlo \cite{https://doi.org/10.48550/arxiv.1402.4102, https://doi.org/10.48550/arxiv.1701.02434, Zhang_Li_Shen_Xie_Qian_2021}, Markov Chain Monte Carlo (MCMC) \cite{brooks2011handbook,10.5555/3104482.3104568, 10.5555/3327757.3327920}, variational inference \cite{NIPS2011_7eb3c8be}, deep ensembles \cite{10.5555/3295222.3295387}, and expectation backpropagation \cite{10.5555/2968826.2968934}.

\noindent\textbf{Variational Inference} Variational inference involves approximating an intractable posterior ($p(\theta|x,y)$) with a tractable distribution $q_{\phi}(\theta)$, where $\phi$ refers to the variational parameters. In the context of Bayesian neural networks, for $q_{\phi}(\theta)$ is defined as a product of independent Gaussian distributions, each corresponding to a single parameter in the network:
\begin{equation}
    q_{\phi}(\theta) = \prod_{j = 1}^{M} \mathcal{N}(\mu_{j}, \sigma_{j}^{2})
\end{equation} 
where $M$ refers to the number of Bayesian weights in the network. This formulation is known as mean-field variational inference. The variational parameters $\phi$ are optimized by maximizing the evidence lower bound (ELBO):
\begin{equation}
    \mathcal{L} = E_{q_{\phi}(\theta)}[log(p(y|\theta,x))] -KL[q_{\phi}(\theta)||p(\theta)] 
\end{equation} 
The first term refers to the expected log likelihood, and the
second term is the Kullback-Leibler (KL) divergence which measures how close $q_{\phi}(\theta)$ is to the prior $p(\theta)$. A popular variational inference approach is the Bayes by Backprop framework \cite{10.5555/3045118.3045290}, which involves optimizing variational parameters by backpropagation. Other approaches involve approximations via either monte-carlo dropout \cite{pmlr-v48-gal16}, the Adam optimizer \cite{pmlr-v80-khan18a}, or multiplicative noise \cite{10.5555/3305890.3305910}.

\noindent\textbf{Empirical Bayes} From a Bayesian viewpoint, priors should accurately reflect our beliefs about the network's parameters $\theta$ before any data is observed. However, Empirical Bayes approaches estimate the prior distribution from data \cite{10.1214/aoms/1177703729}. As mentioned in the introduction, the MOPED framework \cite{Krishnan_Subedar_Tickoo_2020} is an Empirical Bayes, or more specifically, a Parametric Empirical Bayes (PEB) framework, designed to inform parameter priors and posteriors with their Maximum Likelihood Estimate (MLE). In MOPED, the MLE is employed to determine the prior's mean and both the mean and variance of the posterior. This approach has been applied in the fields of systems medicine \cite{Klebanov2016EmpiricalBM} and risk assessment \cite{GRIBOK2020106805}. In our paper, we introduce a PEB approach for open-domain dialogue. We base the prior and approximate posterior parameters on their position in addition to their corresponding MLE. In our case, the MLE is the pretrained parameters in GPT-2/DialoGPT: $\hat{\theta} = \argmin_{\hat{\theta}} \mathcal{L}(\hat{\theta})$, where $\mathcal{L}$ refers to the cross-entropy loss used during GPT-2/DialoGPT pretraining.

\noindent \section{Bayesian Open-Domain Dialogue via Empirical Bayes (BODEB)}
BODEB involves constructing a Bayesian open-domain dialogue agent by leveraging pretrained langauge model parameters for prior definition and approximate posterior initialization. For this paper, we will utilize the GPT-2 and DialoGPT PLMs. While both pretrained models are architecturally identical, they differ when it comes to pretraining. DialoGPT has been pretrained exclusively for the task of multi-turn response generation, while GPT-2's pretraining is more general in scope.

\begin{figure}
    \centering
    \scalebox{0.5}{
    \includegraphics{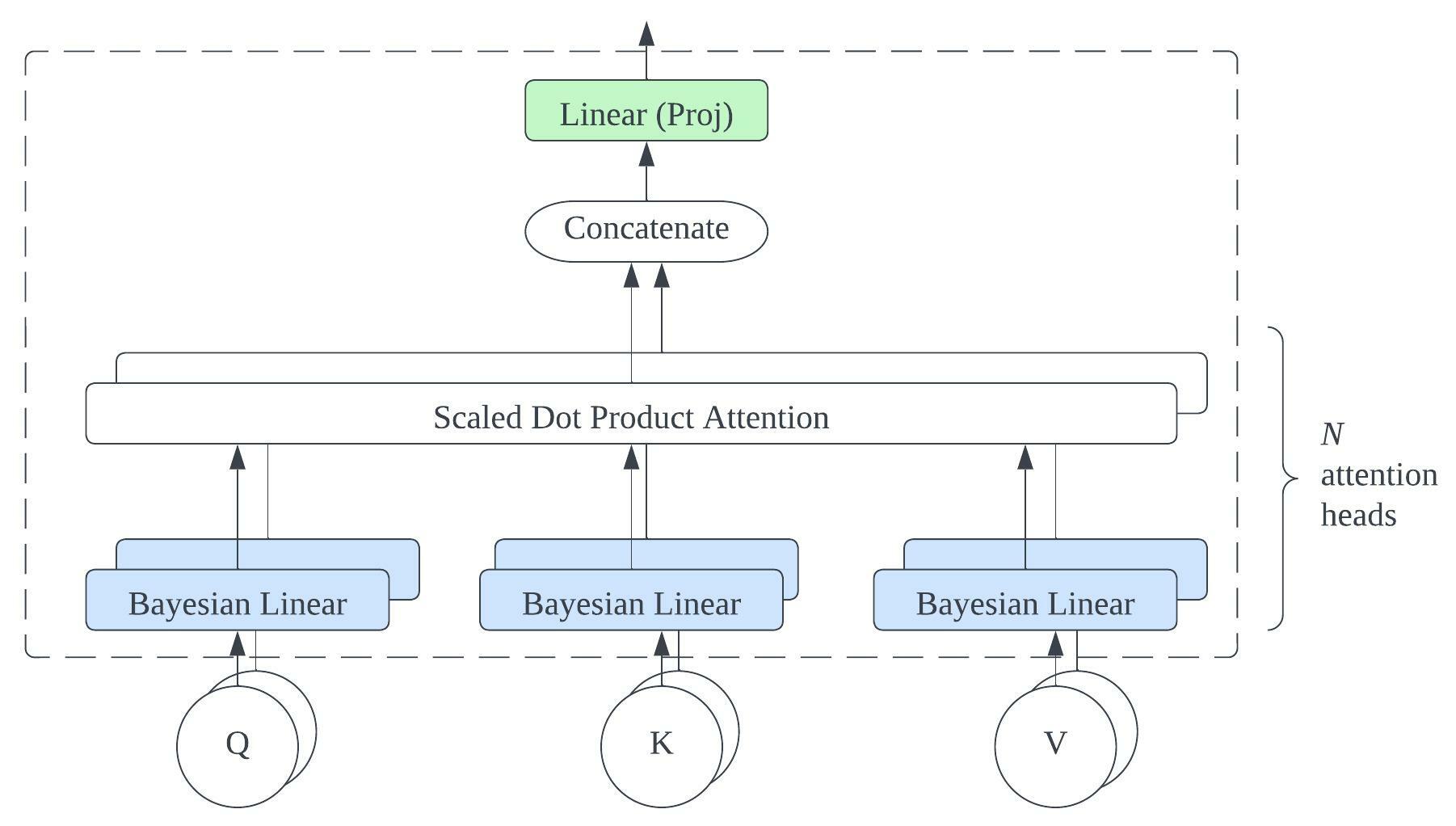}}
    \caption{Self-attention component of the transformer under the BODEB framework. Only attention layers (in blue) are formulated as Bayesian layers. The final linear layer, which we term the projection layer (in green), is deterministic. For GPT-2/DialoGPT, $N=12$.}
    \label{fig:my_label}
\end{figure}
\begin{figure}
    \centering
    \scalebox{0.5}{
    \includegraphics{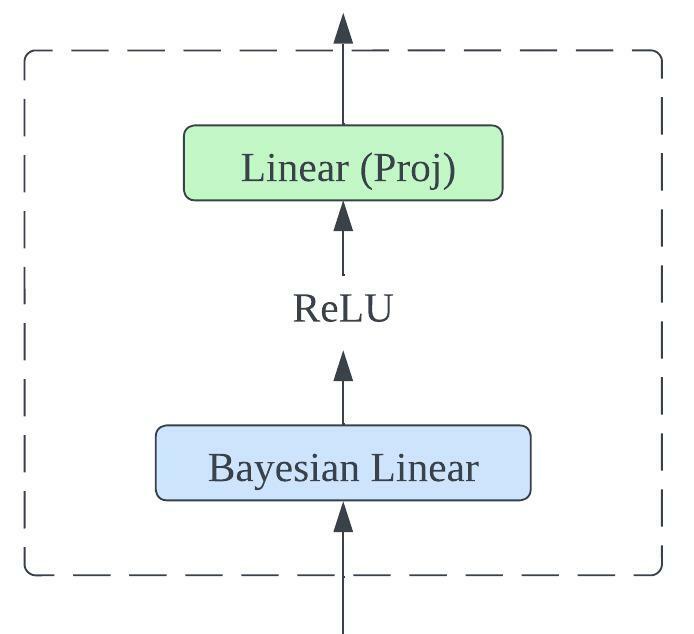}}
    \caption{Feed forward component of the transformer under the BODEB framework. Only the first linear layer (in blue) is formulated as a Bayesian layer. The second linear layer, which we term the projection layer (in green), is deterministic.}
    \label{fig:my_label}
\end{figure}

\subsection{Architecture}

We aim to construct a Bayesian open-domain dialogue agent that is architecturally identical to GPT-2/DialoGPT. Only \emph{selected layers} from the masked self-attention and feed forward components of the transformer decoder are formulated as Bayesian layers. For the masked self-attention component (Figure 1), only the self-attention layers responsible for generating the Query, Key, and Value matrices are formulated as Bayesian layers. For the feed forward component (Figure 2), only the \emph{first} linear layer is formulated as a Bayesian layer. The final linear layer in both the masked self-attention and feed forward component, which we term \emph{projection} layers, are not treated as Bayesian layers, and consist of deterministic parameters. We found that utilizing Bayesian projection layers would adversely affect model performance in terms of coherence in addition to substantially increasing model size.


\subsection{Approximate Posterior}

In our approach, all Bayesian layers employ a posterior distribution approximation, which takes the form of a Gaussian distribution characterized by a mean $\mu$ and a standard deviation $\sigma$. Following the methodology outlined in previous works such as \citet{10.5555/3045118.3045290} and \citet{Krishnan_Subedar_Tickoo_2020}, we adopt the softplus function to ensure that the standard deviation remains non-negative. Thus, we parameterize $\sigma$ as $\sigma = \log(1 + e^{\rho})$. During the fine-tuning process, both the mean $\mu$ and the parameter $\rho$ associated with each Bayesian parameter will be updated iteratively.


The initialization of the mean of the posterior Gaussian distribution is based on the corresponding weight or bias value in the PLM, denoted as $\hat{w}$ and $\hat{b}$, respectively. To determine the standard deviation, we introduce a variable, denoted as $pos$, which signifies the order or position of the parameter within the transformer module in the PLM. For instance, a parameter from the 5th decoder module in GPT-2 will have $pos = 5$. Enforcing constraints on the variance of the Gaussian parameter distributions is crucial to prevent the generation of parameters that exhibit excessive deviations from the mean during inference.

Our hypothesis posits that imposing constraints on the variance, particularly for deeper parameters (i.e., those with larger $pos$ values), will enhance coherence in generated responses while preserving response diversity. For each parameter, the standard deviation $\sigma$ is initialized as the product of three factors: the absolute value of the corresponding pretrained parameter, the position of the module $pos$, and a predefined penalty term denoted as $\alpha$. Consequently, for any weight $w$ and bias $b$ within the $i$th module, the initialization of the approximate posterior Gaussian is as follows:
\begin{equation}
    w \sim \mathcal{N}(\hat{w}, (|\hat{w}|*\frac{1}{pos_{i}} \alpha)^{2})
\end{equation}
\begin{equation}
    \rho_{i} = log(e^{|\hat{w}|*\frac{1}{pos_{i}} \alpha} - 1)
\end{equation}
\begin{equation}
    b \sim \mathcal{N}(\hat{b}, (|\hat{b}|*\frac{1}{pos_{i}} \alpha)^{2})
\end{equation}
\begin{equation}
    \rho_{i} = log(e^{|\hat{b}|*\frac{1}{pos_{i}} \alpha} - 1)
\end{equation}
where $\alpha$ is a hyperparameter to be tuned during finetuning. Increasing $\alpha$ would result in a larger variance initialization and vice versa.

\subsection{Prior}
For the prior, we implement two different priors: a standard Gaussian prior and a mixture prior. 

\noindent \textbf{Gaussian Prior} The Gaussian prior consists of a standard Gaussian distribution similar to the MOPED framework. Similar to the approximate posterior, the softplus function is applied to the standard deviation, which is parameterized as $log(1 + e^{\rho})$. For an arbitrary weight $w$ and bias $b$ in the $i$th transformer module, the prior is defined as follows:
\begin{equation}
    w \sim \mathcal{N}(\hat{w}, (log(1+e^{\rho_{i}}))^{2})
\end{equation}
\begin{equation}
    b \sim \mathcal{N}(\hat{b}, (log(1+e^{\rho_{i}}))^{2})
\end{equation}
\begin{equation}
    \rho_{i} = \frac{1}{pos_{i}}
\end{equation}
where $\hat{w}$ and $\hat{b}$ refer to the value of the corresponding parameter in either GPT-2 or DialoGPT, and $pos_{i}$ represents the position of the $i$th module. $pos_{i}$ ensures that the variance of the prior decreases for deeper parameters. 

\noindent \textbf{Mixture Prior} To impose a tighter constrain on the resultant posterior, we propose a mixture prior consisting of two distinct Gaussians (similar to the original Bayes by Backprop). The mixture prior can be expressed via the following expression:
\begin{equation}
    p(\theta) = \prod_{j = 1}^{M} \eta \mathcal{N}(\mu_{1,j}, \sigma_{1,j}^{2}) + (1-\eta)\mathcal{N}(\mu_{2,j}, \sigma_{2,j}^{2})
\end{equation} 
where the parameter $\eta$ is a tunable hyperparameter that determines the contribution of each Gaussian component, and $\mu_{1,j}, \sigma_{1,j}^{2}$ as well as $\mu_{2,j}, \sigma_{2,j}^{2}$ represent the means and variances defining the first and second Gaussian components. Similar to the approach used for the approximate posterior, we parameterize $\sigma_{1,j}$ and $\sigma_{2,j}$ as $log(1 + e^{\rho_{1}})$ and $log(1 + e^{\rho_{2}})$, respectively.

For both the first and second Gaussian components, the means $\mu_{1,j}$ and $\mu_{2,j}$ are kept fixed at the Maximum Likelihood Estimate (MLE) of the corresponding parameter, and the mean of the resulting mixture Gaussian prior corresponds to either $\hat{w}$ or $\hat{b}$. However, there is a difference in the standard deviation values between the two Gaussians. In the case of the first Gaussian, we set $\rho = 1$. For the second Gaussian, at module $i$, $\rho$ is defined as the inverse of $pos_{i}^{2}$. Typically, the variance of the resulting mixture of two Gaussians is calculated as $\sigma^{2} = \eta\sigma_{1}^{2} + (1-\eta)\sigma_{2}^{2} + \eta(1-\eta)(\mu_{1}-\mu_{2})^{2}$. In our specific case, the third term, which accounts for the shift from the individual means relative to the mixture mean, can be omitted since both Gaussians have identical means. The final mixture Gaussian for any weight $w$ and bias $b$ in module $i$ is then defined as:
\begin{equation}
    w \sim \mathcal{N}(\hat{w}, \eta\sigma_{1}^{2} + (1-\eta)\sigma_{2}^{2})
\end{equation}
\begin{equation}
    b \sim \mathcal{N}(\hat{b}, \eta\sigma_{1}^{2} + (1-\eta)\sigma_{2}^{2})
\end{equation}
\begin{equation}
    \sigma_{1} = log(1+e^{\rho_{1}}); \sigma_{2} = log(1+e^{\rho_{2}})
\end{equation}
\begin{equation}
    \rho_{1} = 1; \rho_{2} = \frac{1}{pos_{i}^{2}}
\end{equation}

Since the variance of the second Gaussian would naturally be much smaller than the variance of the first Gaussian ($\sigma_{2}^{2} << \sigma_{1}^{2} $), our mixture prior would resemble a spike-and-slab prior. Similarly, the $pos_{i}^{2}$ term ensures that the variance of the second Gaussian decreases exponentially for deeper parameters, which emphasizes the spike in the prior distributions for deeper parameters. Thus, unlike the Gaussian prior, the mixture prior encourages the approximate posterior to adopt a spike-and-slab distribution shape during finetuning. This reduces the probability of sampling a parameter that deviates too far from the pretrained parameter value, thereby reducing possibility of generating incoherent responses.

\subsection{Finetuning \& Inference}
Once the Bayesian model is constructed, the model is finetuned on the dialogue corpus. The loss is computed via Equation 3. Deterministic parameters are optimized via standard backpropagation. Bayesian parameters ($\mu,\rho$) are optimized via the reparameterization trick \cite{https://doi.org/10.48550/arxiv.1506.02557}. During inference, the approximated posterior corresponding to every Bayesian parameter is randomly sampled every time a new dialogue context is presented to the model. 

\section{Experiment}

\noindent\textbf{Corpora} For our experiments, we utilize the DailyDialog \cite{li-etal-2017-dailydialog} and EmpatheticDialogs \cite{rashkin-etal-2019-towards} corpora. Both corpora consists of multi-turn conversations between two interlocutors, covering a range of subjects and emotions. Further details can be found in the Appendix (A.1).

\noindent\textbf{Implementation} We implement two variants of the BODEB framework: BODEB\textsubscript{G} and BODEB\textsubscript{M} which utilize the Gaussain prior and mixture prior respectively. Both BODEB\textsubscript{G} and BODEB\textsubscript{M} are implemented with the small version of GPT-2 and DialoGPT provided by HuggingFace, which consists of 12 transformer decoder components ($\sim$ 124 million parameters).  Hence, positional parameter $pos = \{1,2, \cdots, 12\}$. Once the prior and approximate posterior distributions are defined, the Bayesian model is finetuned for three epochs. The AdamW optimizer \cite{https://doi.org/10.48550/arxiv.1711.05101} is used during finetuning (learning rate = 2e-5, batch size = 16). $\alpha$ is fixed at 5e-2. Also, we use greedy decoding to generate all responses. Although decoding strategies such as beam search, random sampling with temperature, as well as top-p and top-k sampling are known to impact response diversity, we use greedy decoding so that any improvements in diversity can be directly attributed to the model.

\noindent\textbf{Baselines} In our study, we assess various Transformer-based and Pretrained Language Model (PLM)-based baselines. Among the Transformer-based models, we train the following architectures: a standard Transformer, a Transformer decoder-based Conditional Variational Autoencoder (CVAE) \cite{zhao-etal-2017-learning}, the Sequential Variational Transformer (SVT) \cite{https://doi.org/10.48550/arxiv.2003.12738}, and the Randomized Link (RL) Transformer \cite{lee-etal-2022-randomized}. The SVT includes a variational decoder layer that generates distinct latent variables for each position, while the RL Transformer introduces stochasticity during inference through additional randomized weights. All Transformer architectures in our experiments consist of four encoders and four decoders. For the PLM-based baselines, we employ GPT-2 and DialoGPT. In addition to fine-tuning these PLMs on dialogue corpora, we implement the following models: a CVAE with a GPT-2/DialoGPT decoder, and the Uncertainty-Aware (UA) CVAE as described in Section 2. Furthermore, we introduce a Bayesian model using the MOPED framework. In all CVAE-based approaches, the latent variable sampled from the prior Gaussian or the approximated posterior (defined by three-layer Multi-Layer Perceptrons) is combined with the input to the decoder for text generation.

\noindent\textbf{Automatic Evaluation} To measure diversity, we utilize the inter-response Distinct-$1$ and $2$ scores \cite{li-etal-2016-diversity}. We also employ traditional lexical diversity metrics such as the Textual Lexical Diversity (MTLD) \cite{fergadiotis2011modeling}, the Moving-Average Type–Token Ratio (MATTR) \cite{covington2007mattr}, and the Hypergeometric Distribution Diversity (HDD) index \cite{mccarthy2007vocd} from the field of linguistics. These metrics effectively measure utterance level diversity. For coherence, \citet{9747458} introduced the Utterance Entailment (UE) score, which involves applying a BERT-based Natural Language Inference (NLI) model to the generated response and each utterance in the dialogue context. For our evaluation, we will implement an improved version of the UE-score which provides a more accurate score in the presence of long multi-utterance contexts (implementation details provided in Appendix A.2.2). Further details can be found in the Appendix (A.2).

Additionally, we do not use automatic metrics drawn from machine translation such as the BLEU \cite{10.3115/1073083.1073135}, ROUGE\cite{lin-2004-rouge}, and METEOR \cite{banerjee-lavie-2005-meteor} scores. Due to the one-to-many property of dialogue (each dialogue context has multiple plausible responses), metrics which measure the similarity of the generated response to the reference response are unsuitable for the task of open-domain dialogue \cite{liu-etal-2016-evaluate, lee-etal-2022-randomized}.

\noindent\textbf{Human Evaluation} We also utilize human evaluation to evaluate the responses generated by the DialoGPT baselines on the DailyDialog corpus. We invited five native English speakers to compare responses based on `Diversity', `Fluency', and `Coherence'. `Diversity' refers to the variability of the generated responses in terms of vocabulary i.e., intra-response word-level diversity, `Fluency' accounts for the eloquence of the responses, and `Coherence' refers to the relevance of each response with regard to the dialogue context. Further details can be found in the Appendix (A.2.4).

\section{Results \& Discussion}
The automatic evaluation and human evaluation results are presented in Table 1 and 3 respectively. Samples of responses generated by DialoGPT-based models are provided in the Appendix (A.3). We also present additional experimental results with different variance configurations in Appendix A.4, and comparisons with different temperature values in Appendix A.5.

Responses generated by PLM-based approaches are far more diverse and coherent relative to transformer-based approaches. GPT-2/DialoGPT approaches attained significantly higher diversity and coherence scores on both corpora. This falls within expectation as PLM-based approaches would naturally possess greater overall language understanding and generation capabilities due to pretraining. Additionally, based on the noticeably higher UE-scores attained, we can conclude that responses generated by DialoGPT-based approaches achieve better performance compared to their GPT-2 counterparts when it comes to coherence. By examining the generated responses, it is apparent that DialoGPT-based responses generally display far more relevance and consistency with respect to the dialogue context. This is also expected as DialoGPT is pretrained specifically for the task of dialogue generation.
\begin{table}[]
\caption{Automatic evaluation results on DailyDialog and EmpatheticDialogs. The best score generated by each PLM baseline is \textbf{bolded}. * indicates statistically significant
differences (t-test, $p$-value \textless 0.01) from the \textbf{bolded} result.}
\centering
\scalebox{0.65}{
\begin{tabular}{llllllllll}
\hline
            & \multicolumn{6}{c}{DailyDialog}                                                                          \\ \cline{2-7} 
            & Dist-1 & Dist-2 & MATTR & MTLD & HDD & UE\\ \hline
Transformer &0.004   &0.010   &0.366  &12.792&0.269&0.032    \\
-GVT        &0.025   &0.161   &0.597  &34.946&0.523&0.025    \\
-SVT        &0.024   &0.152   &0.452  &20.396&0.453&0.011     \\
-RL         &0.043   &0.179   &0.578  &33.261&0.512&0.026    \\ \hline
GPT-2       &0.036*  &0.158*   &0.583*  &23.938*&0.638*&0.094*     \\
-CVAE       &0.048   &0.195*   &0.604* &24.995**&0.652**&0.089*     \\
-UA-CVAE    &0.045*  &0.187*  &0.609* &24.523**&0.644**&0.106*     \\
-BODEB\textsubscript{G}  &0.049   &0.215   &0.625  &27.523&0.665&0.146       \\
-BODEB\textsubscript{M}  &\textbf{0.050}   &\textbf{0.228}   &\textbf{0.635}  &\textbf{29.461}&\textbf{0.696}&\textbf{0.152}           \\ \hline
DialoGPT    &0.043*   &0.207*   &0.653*  &31.547&*0.694*&0.233*      \\
-CVAE       &0.047*   &0.258*   &0.686*  &37.821*&0.715*&0.201*      \\
-UA-CVAE    &0.045*   &0.221*   &0.677*  &35.527*&0.689*&0.215*     \\
-BODEB\textsubscript{G}  &0.050*   &0.323*   &0.718  &47.015&0.743&0.226*    \\
-BODEB\textsubscript{M}  &\textbf{0.056}   &\textbf{0.369}   &\textbf{0.748}  &\textbf{48.949}&\textbf{0.769}&\textbf{0.245}    \\ \hline
            & \multicolumn{6}{c}{EmpatheticDialogues}                                                                  \\ \cline{2-7} 
            & Dist-1 & Dist-2 & MATTR & MTLD & HDD & UE\\ \hline
Transformer &0.012   &0.069   &0.399  &17.562&0.301&0.025     \\
-GVT        &0.035   &0.255   &0.565  &27.364&0.633&0.027     \\
-SVT        &0.029   &0.209   &0.486  &25.675&0.592&0.021     \\
-RL         &0.040   &0.307   &0.606  &29.496&0.622&0.026     \\ \hline
GPT-2       &0.029*   &0.101* &0.454*  &16.466*&0.494*&0.073*      \\
-CVAE       &0.057*   &0.203* &0.547*  &21.289*&0.592*&0.097*     \\
-UA-CVAE    &0.055*   &0.186* &0.521*  &20.342*&0.553*&0.092*     \\
-BODEB\textsubscript{G}  &0.061   &0.236   &\textbf{0.611}  &\textbf{27.651}&\textbf{0.663}&0.101*    \\
-BODEB\textsubscript{M}  &\textbf{0.063}   &\textbf{0.245}   &0.610  &27.054&0.651&\textbf{0.110}     \\ \hline
DialoGPT    &0.049*   &0.211*   &0.615*  &26.466*&0.653*&0.244     \\
-CVAE       &0.048*   &0.263*   &0.607*  &29.791*&0.666*&0.212*     \\
-UA-CVAE    &0.051*   &0.251*   &0.624*  &31.294*&0.685*&0.226*     \\
-BODEB\textsubscript{G}  &0.056   &0.306   &0.688  &37.356&0.729&0.241    \\
-BODEB\textsubscript{M}  &\textbf{0.058}   &\textbf{0.310}   &\textbf{0.700}  &\textbf{39.219}&\textbf{0.731}&\textbf{0.250}  \\ \hline
\end{tabular}}
\end{table}

It is also apparent that responses generated by BODEB demonstrate greater contextual coherence relative to all other baselines. Since both BODEB\textsubscript{G} and BODEB\textsubscript{M} attained higher UE-scores compared to other baselines on both corpora. Furthermore, for human evaluation, BODEB\textsubscript{M} attained a large percentage of wins against DialoGPT, UA-CVAE, and CVAE. BODEB\textsubscript{M} also generally achieves better results in terms of coherence relative to BODEB\textsubscript{G}. This is evidenced by the higher UE-scores attained by BODEB\textsubscript{M} when applied to GPT-2/DialoGPT on both corpora. Human evaluation also supports this observation as BODEB\textsubscript{M} achieves a high percentage of wins and a low percentage of losses when compared to BODEB\textsubscript{G}. This confirms our hypothesis in section 3.2.

\begin{table}[]
\caption{Automatic evaluation results for MOPED and BODEB. The highest score generated by each PLM baseline is \textbf{bolded}.}
\centering
\scalebox{0.65}{
\begin{tabular}{llllllllll}
\hline
            & \multicolumn{6}{c}{DailyDialog}                                                                          \\ \cline{2-7} 
            & Dist-1 & Dist-2 & MATTR & MTLD & HDD & UE\\ \hline
GPT-2       &   &   &  & &&    \\
-MOPED     &\textbf{0.074}   &\textbf{0.296}   &0.375*  &11.371* &0.396*  &0.004*    \\
-BODEB\textsubscript{G}  &0.049*   &0.215*   &0.625  &27.523&0.665&0.146       \\
-BODEB\textsubscript{M}  &0.050*   &0.228*   &\textbf{0.635}  &\textbf{29.461} &\textbf{0.696}&\textbf{0.152}           \\ \hline
DialoGPT    &   &   &  &&&   \\
-MOPED*     &\textbf{0.099}   &\textbf{0.495}   &0.530*  &17.609* &0.562*&0.006*   \\
-BODEB\textsubscript{G}  &0.050*   &0.323*   &0.718  &47.015&0.743&0.226*    \\
-BODEB\textsubscript{M}  &0.056*   &0.369*   &\textbf{0.748}  &\textbf{48.949}&\textbf{0.769}&\textbf{0.245}    \\ \hline
            & \multicolumn{6}{c}{EmpatheticDialogues}                                                                  \\ \cline{2-7} 
            & Dist-1 & Dist-2 & MATTR & MTLD & HDD & UE\\ \hline
GPT-2       &  &   &  &&&     \\
-MOPED     &\textbf{0.079}   &\textbf{0.265}   &0.392*  &12.381* &0.422*  &0.007*    \\
-BODEB\textsubscript{G}  &0.061*   &0.236*   &\textbf{0.611}  &\textbf{27.651}&\textbf{0.663}&0.101    \\
-BODEB\textsubscript{M}  &0.063*   &0.245*   &0.610  &27.054&0.651&\textbf{0.110}     \\ \hline
DialoGPT    &   &   &  &&&    \\

-MOPED     &\textbf{0.094}   &\textbf{0.428}   &0.489*  &15.638* &0.518*&0.010*     \\
-BODEB\textsubscript{G}  &0.056*   &0.306*   &0.688  &37.356 &0.729&0.241    \\
-BODEB\textsubscript{M}  &0.058*   &0.310*   &\textbf{0.700}  &\textbf{39.219} &\textbf{0.731}&\textbf{0.250}  \\ \hline
\end{tabular}}
\end{table}
\begin{figure}
    \centering
    \scalebox{0.25}{
    \includegraphics{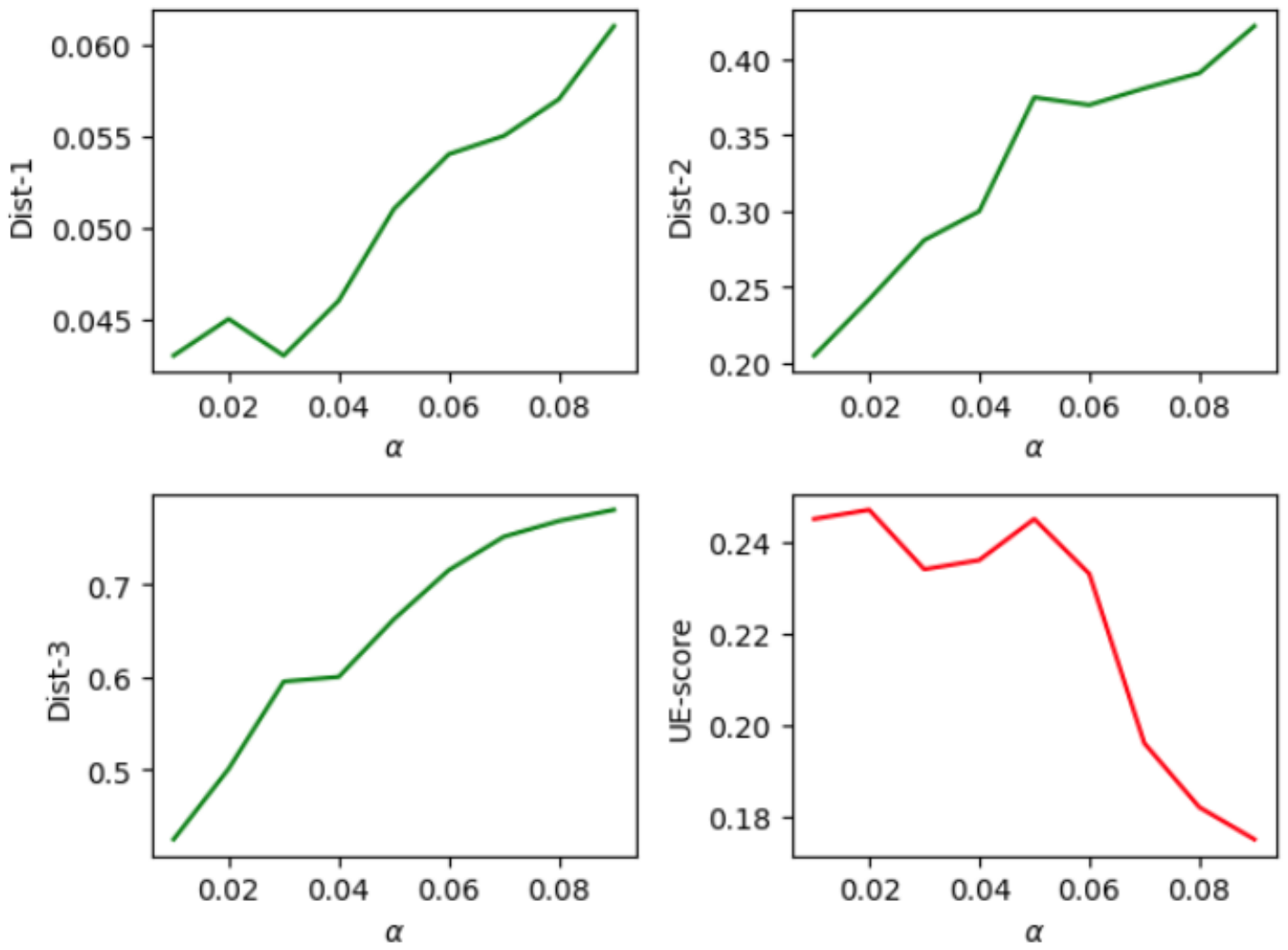}
    }
    \caption{Line plots depicting the relationship between $\alpha$ and the distinct-1,2,  3 scores as well as UE-score (clockwise from top left) for BODEB\textsubscript{M}.}
    \label{fig:my_label}
\end{figure}

\begin{table*}
\centering
\caption{Human evaluation results for DialoGPT on the DailyDialog corpus. `W',`T', and `L' represent the percentage of Wins, Ties and Losses respectively. The Kappa scores ($\kappa$) \cite{fleiss1971mns} generally range from 0.5 to 0.6 indicating moderate inter-rater agreement.}
\scalebox{0.68}{
\begin{tabular}{llllll} 
\hline
          & \multicolumn{1}{c}{\begin{tabular}[c]{@{}c@{}}BODEB\textsubscript{M} vs DialoGPT\\ (W/T/L/$\kappa$)\end{tabular}} & \multicolumn{1}{c}{\begin{tabular}[c]{@{}c@{}}BODEB\textsubscript{M} vs CVAE\\ (W/T/L/$\kappa$)\end{tabular}} & \multicolumn{1}{c}{\begin{tabular}[c]{@{}c@{}}BODEB\textsubscript{M} vs UA-CVAE\\ (W/T/L/$\kappa$)\end{tabular}} & \multicolumn{1}{c}{\begin{tabular}[c]{@{}c@{}}BODEB\textsubscript{M} vs. MOPED\\ (W/T/L/$\kappa$)\end{tabular}} & \multicolumn{1}{c}{\begin{tabular}[c]{@{}c@{}}BODEB\textsubscript{M} vs. BODEB\textsubscript{G}\\(W/T/L/\textbackslash{}$\kappa$)\end{tabular}}  \\ 
\hline
Diversity & 65\%/21\%/14\%/0.63                                                                              & 51\%/28\%/21\%/0.56                                                                          & 55\%/30\%/15\%/0.51                                                                             & 40\%/31\%/29\%/0.56                                                                            & 41\%/33\%/26\%/0.54                                                                                         \\
Fluency   & 40\%/45\%/15\%/0.56                                                                              & 46\%/45\%/9\%/0.52                                                                          & 44\%/39\%/17\%/0.65                                                                             & 72\%/22\%/6\%/0.61                                                                            & 39\%/47\%/14\%/0.52                                                                                         \\
Coherence & 45\%/44\%/11\%/0.53                                                                              & 51\%/37\%/12\%/0.58                                                                          & 53\%/39\%/8\%/0.52                                                                             & 81\%/12\%/7\%/0.70                                                                            & 47\%/42\%/11\%/0.55                                                                                         \\
\hline
\end{tabular}}
\end{table*}

\noindent \textbf{Comparison with MOPED} We directly apply MOPED to GPT-2 and DialoGPT and summarized the evaluation results in Table 2. It's important to note that MOPED-generated responses were mostly \emph{nonsensical, incoherent gibberish}, which led to the high Distinct-1 and 2 scores and very low UE scores in Table 2. Examples provided in Appendix A.3. Furthermore, in Table 3, when it comes to Fluency and Coherence, BODEB\textsubscript{M} attained a vast majority of wins, emphasizing the nonsensical, incoherent nature of MOPED-generated responses.

\noindent \textbf{Ablation Study} 
We conducted an ablation study on BODEB\textsubscript{M} to assess the impact of each Bayesian layer on overall performance in Table 4. Our findings indicate that treating the projection layers as Bayesian has a detrimental effect on the dialogue agent's comprehension and generation abilities. This leads to the generation of incoherent and nonsensical responses, as evident from the low coherence and high diversity scores. Furthermore, in comparison to the self-attention layer (\emph{Attn}) and the initial linear layer in the feed-forward network (\emph{FF}), incorporating a Bayesian language modeling output layer (\emph{LM Head}) yields the most substantial improvement in diversity, as indicated by the significant increase in diversity scores when the language modeling head is implemented as a Bayesian layer.

\begin{table}[]
\caption{Automatic evaluation results of the ablation study for BODEB\textsubscript{M} (DialoGPT) on the DailyDialog corpus. `-' and `+' denotes a deterministic layer and Bayesian layer respectively. \emph{Attn}, \emph{FC}, \emph{LM Head}, and \emph{Proj} refer to the self-attention layers in the masked self-attention component, the first linear layer in the feed-forward component, the output language modelling head, and the projection layers in both the masked self-attention and feed forward components respectively.}
\centering
\scalebox{0.65}{
\begin{tabular}{llllllllll}
\hline
            & Dist-1 & Dist-2 & MATTR & MTLD & HDD & UE\\ \hline
BODEB\textsubscript{M}       &0.056   &0.369      &0.748  &\textbf{48.949}&\textbf{0.769}&\textbf{0.245}    \\
-\emph{Attn}                 &0.046*   &0.286*      &0.668*       &46.523*      &0.702*     &0.232*         \\
-\emph{FC}                   &0.047*   &0.274*      &0.659*       &45.974*      &0.684*     &0.234*          \\
-\emph{LM Head}              &0.038*   &0.223*      &0.602*       &41.269*      &0.657*     &0.239*         \\
+\emph{Proj}                 &\textbf{0.058}   &\textbf{0.381}      &\textbf{0.776}       &48.237      &0.753     &0.196*         \\ \hline
\end{tabular}}
\end{table}

\noindent \textbf{Impact of Variance on Diversity \& Coherence} 
The variance of parameter Gaussians significantly affects overall performance. Specifically, larger variances in parameter distributions tend to enhance response diversity but diminish contextual coherence. In the case of BODEB, the variance in the approximate posterior mainly depends on the hyperparameter $\alpha$. We illustrate the relationship between $\alpha$ and distinct-1,2,3 scores as well as UE-score for BODEB\textsubscript{M} in Figure 3. We propose that a substantial parameter variance ($\alpha > $ 5e-2) increases the likelihood of randomly selecting a weight or bias that deviates significantly from the mean. This undermines the advantages of using pretrained parameters, leading to a decline in the model's language understanding and generation capabilities, resulting in gibberish generation, which in turn implies higher distinct scores and lower UE scores.

\noindent \textbf{Application in Large Language Models (LLMs)} BODEB can also be applied directly to open-source LLMs such as Falcon or Vicuna. However, due to computational resource limitations, we could not apply BODEB to these LLMs, or compare its performance to state-of-the-art variants such as GPT-4. Similar to our findings with GPT-2/DialoGPT, we expect that applying BODEB directly to LLMs will enhance response diversity while maintaining coherence. However, we further posit that performance of the resultant Bayesian LLM could be enhanced through additional fine-tuning or instruction tuning after BODEB is applied (Section 3.4). Investigating the efficacy of BODEB in the context of LLMs represents a promising avenue for future research. 

It should be noted that applying the BODEB framework would entail a relatively large increase in model size as selected layers in the model are formulated as Bayesian layers. Additional fine-tuning or instruction-tuning would also require substantial computational resources. This issue could be potentially mitigated by only applying BODEB to specific transformer components in the LLM, which are selected with a certain probability $p$ from a Bernoulli distribution ($p$ is a hyperparameter to be optimized). Further exploration of more strategic selection methods would present a promising avenue for further research.

\section{Conclusion}
In this paper, we introduced BODEB, an Empirical Bayes framework for creating a Bayesian open-domain dialogue agent that can be directly applied to any PLM. We empirically demonstrate that a BODEB-based Bayesian dialogue agent is capable of producing more diverse and coherent responses compared to variational agents. As BNNs allow for predictive uncertainty quantification, future research could involve exploring potential correlations between predictive uncertainty (comprising aleatoric and epistemic uncertainty) and various aspects of response quality, such as diversity and coherence. Another avenue for future work could also involve exploring the efficacy of BODEB when applied to other language generation tasks.

\section{Limitations}
The BODEB framework entails an increase in the total number of parameters, which translates to greater model size. For larger PLMs/LLMs, this could constitute a relatively significant increase in memory requirement. Additionally, for each new dialogue context fed to the Bayesian PLM/LLM, every Bayesian parameter will have to be sampled, resulting in additional latency during inference. Finally, as mentioned in Section 5, due to computational resource limitations, we did not apply BODEB to LLMs such as Falcon or Vicuna, which would have allowed for comparison with state-of-the-art LLMs such as GPT-4. Examining the effectiveness of BODEB in open-source LLMs is a promising direction for future work.

\bibliography{anthology,custom}

\begin{thebibliography}{54}
\expandafter\ifx\csname natexlab\endcsname\relax\def\natexlab#1{#1}\fi

\bibitem[{Banerjee and Lavie(2005)}]{banerjee-lavie-2005-meteor}
Satanjeev Banerjee and Alon Lavie. 2005.
\newblock {METEOR}: An automatic metric for {MT} evaluation with improved
  correlation with human judgments.
\newblock In \emph{Proceedings of the {ACL} Workshop on Intrinsic and Extrinsic
  Evaluation Measures for Machine Translation and/or Summarization}, pages
  65--72, Ann Arbor, Michigan. Association for Computational Linguistics.

\bibitem[{Bao et~al.(2020)Bao, He, Wang, Wu, and Wang}]{bao-etal-2020-plato}
Siqi Bao, Huang He, Fan Wang, Hua Wu, and Haifeng Wang. 2020.
\newblock \href {https://doi.org/10.18653/v1/2020.acl-main.9} {{PLATO}:
  Pre-trained dialogue generation model with discrete latent variable}.
\newblock In \emph{Proceedings of the 58th Annual Meeting of the Association
  for Computational Linguistics}, pages 85--96, Online. Association for
  Computational Linguistics.

\bibitem[{Betancourt(2017)}]{https://doi.org/10.48550/arxiv.1701.02434}
Michael Betancourt. 2017.
\newblock \href {https://doi.org/10.48550/ARXIV.1701.02434} {A conceptual
  introduction to hamiltonian monte carlo}.

\bibitem[{Blundell et~al.(2015)Blundell, Cornebise, Kavukcuoglu, and
  Wierstra}]{10.5555/3045118.3045290}
Charles Blundell, Julien Cornebise, Koray Kavukcuoglu, and Daan Wierstra. 2015.
\newblock Weight uncertainty in neural networks.
\newblock In \emph{Proceedings of the 32nd International Conference on
  International Conference on Machine Learning - Volume 37}, ICML'15, page
  1613–1622. JMLR.org.

\bibitem[{Brooks et~al.(2011)Brooks, Gelman, Jones, and
  Meng}]{brooks2011handbook}
Steve Brooks, Andrew Gelman, Galin Jones, and Xiao-Li Meng. 2011.
\newblock \emph{Handbook of Markov Chain Monte Carlo}.
\newblock CRC press.

\bibitem[{Brosse et~al.(2018)Brosse, Moulines, and
  Durmus}]{10.5555/3327757.3327920}
Nicolas Brosse, \'{E}ric Moulines, and Alain Durmus. 2018.
\newblock The promises and pitfalls of stochastic gradient langevin dynamics.
\newblock In \emph{Proceedings of the 32nd International Conference on Neural
  Information Processing Systems}, NIPS'18, page 8278–8288, Red Hook, NY,
  USA. Curran Associates Inc.

\bibitem[{Brown et~al.(2020)Brown, Mann, Ryder, Subbiah, Kaplan, Dhariwal,
  Neelakantan, Shyam, Sastry, Askell, Agarwal, Herbert-Voss, Krueger, Henighan,
  Child, Ramesh, Ziegler, Wu, Winter, Hesse, Chen, Sigler, Litwin, Gray, Chess,
  Clark, Berner, McCandlish, Radford, Sutskever, and
  Amodei}]{https://doi.org/10.48550/arxiv.2005.14165}
Tom~B. Brown, Benjamin Mann, Nick Ryder, Melanie Subbiah, Jared Kaplan,
  Prafulla Dhariwal, Arvind Neelakantan, Pranav Shyam, Girish Sastry, Amanda
  Askell, Sandhini Agarwal, Ariel Herbert-Voss, Gretchen Krueger, Tom Henighan,
  Rewon Child, Aditya Ramesh, Daniel~M. Ziegler, Jeffrey Wu, Clemens Winter,
  Christopher Hesse, Mark Chen, Eric Sigler, Mateusz Litwin, Scott Gray,
  Benjamin Chess, Jack Clark, Christopher Berner, Sam McCandlish, Alec Radford,
  Ilya Sutskever, and Dario Amodei. 2020.
\newblock \href {https://doi.org/10.48550/ARXIV.2005.14165} {Language models
  are few-shot learners}.

\bibitem[{Cai and Cai(2022)}]{ijcai2022p564}
Zefeng Cai and Zerui Cai. 2022.
\newblock \href {https://doi.org/10.24963/ijcai.2022/564} {Pcvae: Generating
  prior context for dialogue response generation}.
\newblock In \emph{Proceedings of the Thirty-First International Joint
  Conference on Artificial Intelligence, {IJCAI-22}}, pages 4065--4071.
  International Joint Conferences on Artificial Intelligence Organization.
\newblock Main Track.

\bibitem[{Chen et~al.(2014)Chen, Fox, and
  Guestrin}]{https://doi.org/10.48550/arxiv.1402.4102}
Tianqi Chen, Emily~B. Fox, and Carlos Guestrin. 2014.
\newblock \href {https://doi.org/10.48550/ARXIV.1402.4102} {Stochastic gradient
  hamiltonian monte carlo}.

\bibitem[{Covington(2007)}]{covington2007mattr}
MA~Covington. 2007.
\newblock Mattr user manual (caspr research report 2007--05).
\newblock \emph{Atheus, GA}.

\bibitem[{Devlin et~al.(2019)Devlin, Chang, Lee, and
  Toutanova}]{devlin-etal-2019-bert}
Jacob Devlin, Ming-Wei Chang, Kenton Lee, and Kristina Toutanova. 2019.
\newblock \href {https://doi.org/10.18653/v1/N19-1423} {{BERT}: Pre-training of
  deep bidirectional transformers for language understanding}.
\newblock In \emph{Proceedings of the 2019 Conference of the North {A}merican
  Chapter of the Association for Computational Linguistics: Human Language
  Technologies, Volume 1 (Long and Short Papers)}, pages 4171--4186,
  Minneapolis, Minnesota. Association for Computational Linguistics.

\bibitem[{Fergadiotis(2011)}]{fergadiotis2011modeling}
Gerasimos Fergadiotis. 2011.
\newblock \emph{Modeling lexical diversity across language sampling and
  estimation techniques}.
\newblock Arizona State University.

\bibitem[{Fleiss et~al.(1971)}]{fleiss1971mns}
J.L. Fleiss et~al. 1971.
\newblock {Measuring nominal scale agreement among many raters}.
\newblock \emph{Psychological Bulletin}, 76(5):378--382.

\bibitem[{Gal and Ghahramani(2016)}]{pmlr-v48-gal16}
Yarin Gal and Zoubin Ghahramani. 2016.
\newblock \href {https://proceedings.mlr.press/v48/gal16.html} {Dropout as a
  bayesian approximation: Representing model uncertainty in deep learning}.
\newblock In \emph{Proceedings of The 33rd International Conference on Machine
  Learning}, volume~48 of \emph{Proceedings of Machine Learning Research},
  pages 1050--1059, New York, New York, USA. PMLR.

\bibitem[{Gao et~al.(2019)Gao, Bi, Liu, Li, Zhou, and
  Shi}]{gao-etal-2019-discrete}
Jun Gao, Wei Bi, Xiaojiang Liu, Junhui Li, Guodong Zhou, and Shuming Shi. 2019.
\newblock A discrete {CVAE} for response generation on short-text conversation.
\newblock In \emph{Proceedings of the 2019 Conference on Empirical Methods in
  Natural Language Processing and the 9th International Joint Conference on
  Natural Language Processing (EMNLP-IJCNLP)}, pages 1898--1908, Hong Kong,
  China. Association for Computational Linguistics.

\bibitem[{Graves(2011)}]{NIPS2011_7eb3c8be}
Alex Graves. 2011.
\newblock \href
  {https://proceedings.neurips.cc/paper/2011/file/7eb3c8be3d411e8ebfab08eba5f49632-Paper.pdf}
  {Practical variational inference for neural networks}.
\newblock In \emph{Advances in Neural Information Processing Systems},
  volume~24. Curran Associates, Inc.

\bibitem[{Gribok et~al.(2020)Gribok, Agarwal, and Yadav}]{GRIBOK2020106805}
Andrei Gribok, Vivek Agarwal, and Vaibhav Yadav. 2020.
\newblock \href {https://doi.org/https://doi.org/10.1016/j.ress.2020.106805}
  {Performance of empirical bayes estimation techniques used in probabilistic
  risk assessment}.
\newblock \emph{Reliability Engineering \& System Safety}, 201:106805.

\bibitem[{Khan et~al.(2018)Khan, Nielsen, Tangkaratt, Lin, Gal, and
  Srivastava}]{pmlr-v80-khan18a}
Mohammad Khan, Didrik Nielsen, Voot Tangkaratt, Wu~Lin, Yarin Gal, and Akash
  Srivastava. 2018.
\newblock \href {https://proceedings.mlr.press/v80/khan18a.html} {Fast and
  scalable {B}ayesian deep learning by weight-perturbation in {A}dam}.
\newblock In \emph{Proceedings of the 35th International Conference on Machine
  Learning}, volume~80 of \emph{Proceedings of Machine Learning Research},
  pages 2611--2620. PMLR.

\bibitem[{Kingma et~al.(2015)Kingma, Salimans, and
  Welling}]{https://doi.org/10.48550/arxiv.1506.02557}
Diederik~P. Kingma, Tim Salimans, and Max Welling. 2015.
\newblock \href {https://doi.org/10.48550/ARXIV.1506.02557} {Variational
  dropout and the local reparameterization trick}.

\bibitem[{Kingma and Welling(2013)}]{https://doi.org/10.48550/arxiv.1312.6114}
Diederik~P Kingma and Max Welling. 2013.
\newblock \href {https://doi.org/10.48550/ARXIV.1312.6114} {Auto-encoding
  variational bayes}.

\bibitem[{Klebanov et~al.(2016)Klebanov, Sikorski, Schutte, and
  Roblitz}]{Klebanov2016EmpiricalBM}
Ilja Klebanov, Alexander Sikorski, Christof Schutte, and Susanna Roblitz. 2016.
\newblock Empirical bayes methods for prior estimation in systems medicine.
\newblock \emph{arXiv: Methodology}.

\bibitem[{Krishnan et~al.(2020)Krishnan, Subedar, and
  Tickoo}]{Krishnan_Subedar_Tickoo_2020}
Ranganath Krishnan, Mahesh Subedar, and Omesh Tickoo. 2020.
\newblock \href {https://doi.org/10.1609/aaai.v34i04.5875} {Specifying weight
  priors in bayesian deep neural networks with empirical bayes}.
\newblock \emph{Proceedings of the AAAI Conference on Artificial Intelligence},
  34(04):4477--4484.

\bibitem[{Lakshminarayanan et~al.(2017)Lakshminarayanan, Pritzel, and
  Blundell}]{10.5555/3295222.3295387}
Balaji Lakshminarayanan, Alexander Pritzel, and Charles Blundell. 2017.
\newblock Simple and scalable predictive uncertainty estimation using deep
  ensembles.
\newblock In \emph{Proceedings of the 31st International Conference on Neural
  Information Processing Systems}, NIPS'17, page 6405–6416, Red Hook, NY,
  USA. Curran Associates Inc.

\bibitem[{Lee et~al.(2022{\natexlab{a}})Lee, Aik~Lee, and Gan}]{9747458}
Jing~Yang Lee, Kong Aik~Lee, and Woon~Seng Gan. 2022{\natexlab{a}}.
\newblock Improving contextual coherence in variational personalized and
  empathetic dialogue agents.
\newblock In \emph{ICASSP 2022 - 2022 IEEE International Conference on
  Acoustics, Speech and Signal Processing (ICASSP)}, pages 7052--7056.

\bibitem[{Lee et~al.(2022{\natexlab{b}})Lee, Lee, and
  Gan}]{lee-etal-2022-randomized}
Jing~Yang Lee, Kong~Aik Lee, and Woon~Seng Gan. 2022{\natexlab{b}}.
\newblock A randomized link transformer for diverse open-domain dialogue
  generation.
\newblock In \emph{Proceedings of the 4th Workshop on NLP for Conversational
  AI}, pages 1--11, Dublin, Ireland. Association for Computational Linguistics.

\bibitem[{Li et~al.(2016)Li, Galley, Brockett, Gao, and
  Dolan}]{li-etal-2016-diversity}
Jiwei Li, Michel Galley, Chris Brockett, Jianfeng Gao, and Bill Dolan. 2016.
\newblock A diversity-promoting objective function for neural conversation
  models.
\newblock In \emph{Proceedings of the 2016 Conference of the North {A}merican
  Chapter of the Association for Computational Linguistics: Human Language
  Technologies}, pages 110--119, San Diego, California. Association for
  Computational Linguistics.

\bibitem[{Li et~al.(2020)Li, Feng, Wang, Song, Zhang, and Wang}]{ijcai2020p503}
Shifeng Li, Shi Feng, Daling Wang, Kaisong Song, Yifei Zhang, and Weichao Wang.
  2020.
\newblock \href {https://doi.org/10.24963/ijcai.2020/503} {Emoelicitor: An open
  domain response generation model with user emotional reaction awareness}.
\newblock In \emph{Proceedings of the Twenty-Ninth International Joint
  Conference on Artificial Intelligence, {IJCAI-20}}, pages 3637--3643.
  International Joint Conferences on Artificial Intelligence Organization.
\newblock Main track.

\bibitem[{Li et~al.(2017)Li, Su, Shen, Li, Cao, and
  Niu}]{li-etal-2017-dailydialog}
Yanran Li, Hui Su, Xiaoyu Shen, Wenjie Li, Ziqiang Cao, and Shuzi Niu. 2017.
\newblock {D}aily{D}ialog: A manually labelled multi-turn dialogue dataset.
\newblock In \emph{Proceedings of the Eighth International Joint Conference on
  Natural Language Processing (Volume 1: Long Papers)}, pages 986--995, Taipei,
  Taiwan. Asian Federation of Natural Language Processing.

\bibitem[{Lin(2004)}]{lin-2004-rouge}
Chin-Yew Lin. 2004.
\newblock {ROUGE}: A package for automatic evaluation of summaries.
\newblock In \emph{Text Summarization Branches Out}, pages 74--81, Barcelona,
  Spain. Association for Computational Linguistics.

\bibitem[{Lin et~al.(2020)Lin, Winata, Xu, Liu, and
  Fung}]{https://doi.org/10.48550/arxiv.2003.12738}
Zhaojiang Lin, Genta~Indra Winata, Peng Xu, Zihan Liu, and Pascale Fung. 2020.
\newblock \href {https://doi.org/10.48550/ARXIV.2003.12738} {Variational
  transformers for diverse response generation}.

\bibitem[{Liu et~al.(2016)Liu, Lowe, Serban, Noseworthy, Charlin, and
  Pineau}]{liu-etal-2016-evaluate}
Chia-Wei Liu, Ryan Lowe, Iulian Serban, Mike Noseworthy, Laurent Charlin, and
  Joelle Pineau. 2016.
\newblock How {NOT} to evaluate your dialogue system: An empirical study of
  unsupervised evaluation metrics for dialogue response generation.
\newblock In \emph{Proceedings of the 2016 Conference on Empirical Methods in
  Natural Language Processing}, pages 2122--2132, Austin, Texas. Association
  for Computational Linguistics.

\bibitem[{Loshchilov and
  Hutter(2017)}]{https://doi.org/10.48550/arxiv.1711.05101}
Ilya Loshchilov and Frank Hutter. 2017.
\newblock \href {https://doi.org/10.48550/ARXIV.1711.05101} {Decoupled weight
  decay regularization}.

\bibitem[{Louizos and Welling(2017)}]{10.5555/3305890.3305910}
Christos Louizos and Max Welling. 2017.
\newblock Multiplicative normalizing flows for variational bayesian neural
  networks.
\newblock In \emph{Proceedings of the 34th International Conference on Machine
  Learning - Volume 70}, ICML'17, page 2218–2227. JMLR.org.

\bibitem[{Luo and Chien(2021)}]{9414586}
Tien-Ching Luo and Jen-Tzung Chien. 2021.
\newblock \href {https://doi.org/10.1109/ICASSP39728.2021.9414586} {Variational
  dialogue generation with normalizing flows}.
\newblock In \emph{ICASSP 2021 - 2021 IEEE International Conference on
  Acoustics, Speech and Signal Processing (ICASSP)}, pages 7778--7782.

\bibitem[{MacKay(1992)}]{journals/neco/MacKay92a}
David J.~C. MacKay. 1992.
\newblock \href
  {http://dblp.uni-trier.de/db/journals/neco/neco4.html#MacKay92a} {A practical
  bayesian framework for backpropagation networks.}
\newblock \emph{Neural Comput.}, 4(3):448--472.

\bibitem[{McCarthy and Jarvis(2007)}]{mccarthy2007vocd}
Philip~M McCarthy and Scott Jarvis. 2007.
\newblock vocd: A theoretical and empirical evaluation.
\newblock \emph{Language Testing}, 24(4):459--488.

\bibitem[{Papineni et~al.(2002)Papineni, Roukos, Ward, and
  Zhu}]{10.3115/1073083.1073135}
Kishore Papineni, Salim Roukos, Todd Ward, and Wei-Jing Zhu. 2002.
\newblock \href {https://doi.org/10.3115/1073083.1073135} {Bleu: A method for
  automatic evaluation of machine translation}.
\newblock In \emph{Proceedings of the 40th Annual Meeting on Association for
  Computational Linguistics}, ACL '02, page 311–318, USA. Association for
  Computational Linguistics.

\bibitem[{Radford et~al.(2019)Radford, Wu, Child, Luan, Amodei, and
  Sutskever}]{radford2019language}
Alec Radford, Jeff Wu, Rewon Child, David Luan, Dario Amodei, and Ilya
  Sutskever. 2019.
\newblock Language models are unsupervised multitask learners.

\bibitem[{Raffel et~al.(2019)Raffel, Shazeer, Roberts, Lee, Narang, Matena,
  Zhou, Li, and Liu}]{https://doi.org/10.48550/arxiv.1910.10683}
Colin Raffel, Noam Shazeer, Adam Roberts, Katherine Lee, Sharan Narang, Michael
  Matena, Yanqi Zhou, Wei Li, and Peter~J. Liu. 2019.
\newblock \href {https://doi.org/10.48550/ARXIV.1910.10683} {Exploring the
  limits of transfer learning with a unified text-to-text transformer}.

\bibitem[{Rashkin et~al.(2019)Rashkin, Smith, Li, and
  Boureau}]{rashkin-etal-2019-towards}
Hannah Rashkin, Eric~Michael Smith, Margaret Li, and Y-Lan Boureau. 2019.
\newblock \href {https://doi.org/10.18653/v1/P19-1534} {Towards empathetic
  open-domain conversation models: A new benchmark and dataset}.
\newblock In \emph{Proceedings of the 57th Annual Meeting of the Association
  for Computational Linguistics}, pages 5370--5381, Florence, Italy.
  Association for Computational Linguistics.

\bibitem[{Robbins(1964)}]{10.1214/aoms/1177703729}
Herbert Robbins. 1964.
\newblock \href {https://doi.org/10.1214/aoms/1177703729} {{The Empirical Bayes
  Approach to Statistical Decision Problems}}.
\newblock \emph{The Annals of Mathematical Statistics}, 35(1):1 -- 20.

\bibitem[{Ruan and Ling(2021)}]{9406340}
Yu-Ping Ruan and Zhenhua Ling. 2021.
\newblock \href {https://doi.org/10.1109/TAFFC.2021.3073809}
  {Emotion-regularized conditional variational autoencoder for emotional
  response generation}.
\newblock \emph{IEEE Transactions on Affective Computing}, pages 1--1.

\bibitem[{Shen et~al.(2017)Shen, Su, Li, Li, Niu, Zhao, Aizawa, and
  Long}]{shen-etal-2017-conditional}
Xiaoyu Shen, Hui Su, Yanran Li, Wenjie Li, Shuzi Niu, Yang Zhao, Akiko Aizawa,
  and Guoping Long. 2017.
\newblock \href {https://doi.org/10.18653/v1/P17-2080} {A conditional
  variational framework for dialog generation}.
\newblock In \emph{Proceedings of the 55th Annual Meeting of the Association
  for Computational Linguistics (Volume 2: Short Papers)}, pages 504--509,
  Vancouver, Canada. Association for Computational Linguistics.

\bibitem[{Sohn et~al.(2015)Sohn, Lee, and Yan}]{NIPS2015_8d55a249}
Kihyuk Sohn, Honglak Lee, and Xinchen Yan. 2015.
\newblock \href
  {https://proceedings.neurips.cc/paper/2015/file/8d55a249e6baa5c06772297520da2051-Paper.pdf}
  {Learning structured output representation using deep conditional generative
  models}.
\newblock In \emph{Advances in Neural Information Processing Systems},
  volume~28. Curran Associates, Inc.

\bibitem[{Soudry et~al.(2014)Soudry, Hubara, and
  Meir}]{10.5555/2968826.2968934}
Daniel Soudry, Itay Hubara, and Ron Meir. 2014.
\newblock Expectation backpropagation: Parameter-free training of multilayer
  neural networks with continuous or discrete weights.
\newblock In \emph{Proceedings of the 27th International Conference on Neural
  Information Processing Systems - Volume 1}, NIPS'14, page 963–971,
  Cambridge, MA, USA. MIT Press.

\bibitem[{Sun et~al.(2021)Sun, Feng, Li, Liu, and
  Li}]{sun-etal-2021-generating}
Bin Sun, Shaoxiong Feng, Yiwei Li, Jiamou Liu, and Kan Li. 2021.
\newblock \href {https://doi.org/10.18653/v1/2021.acl-long.437} {Generating
  relevant and coherent dialogue responses using self-separated conditional
  variational {A}uto{E}ncoders}.
\newblock In \emph{Proceedings of the 59th Annual Meeting of the Association
  for Computational Linguistics and the 11th International Joint Conference on
  Natural Language Processing (Volume 1: Long Papers)}, pages 5624--5637,
  Online. Association for Computational Linguistics.

\bibitem[{Tolstikhin et~al.(2017)Tolstikhin, Bousquet, Gelly, and
  Schoelkopf}]{https://doi.org/10.48550/arxiv.1711.01558}
Ilya Tolstikhin, Olivier Bousquet, Sylvain Gelly, and Bernhard Schoelkopf.
  2017.
\newblock \href {https://doi.org/10.48550/ARXIV.1711.01558} {Wasserstein
  auto-encoders}.

\bibitem[{Wang et~al.(2022)Wang, Liao, Yu, Wang, Zhang, and
  Liu}]{https://doi.org/10.48550/arxiv.2207.12696}
Ye~Wang, Jingbo Liao, Hong Yu, Guoyin Wang, Xiaoxia Zhang, and Li~Liu. 2022.
\newblock \href {https://doi.org/10.48550/ARXIV.2207.12696} {Advanced
  conditional variational autoencoders (a-cvae): Towards interpreting
  open-domain conversation generation via disentangling latent feature
  representation}.

\bibitem[{Welling and Teh(2011)}]{10.5555/3104482.3104568}
Max Welling and Yee~Whye Teh. 2011.
\newblock Bayesian learning via stochastic gradient langevin dynamics.
\newblock In \emph{Proceedings of the 28th International Conference on
  International Conference on Machine Learning}, ICML'11, page 681–688,
  Madison, WI, USA. Omnipress.

\bibitem[{Wu et~al.(2020)Wu, Li, Wang, Chen, Wong, Feng, Huang, and
  Wang}]{wu-etal-2020-guiding}
Bowen Wu, MengYuan Li, Zongsheng Wang, Yifu Chen, Derek~F. Wong, Qihang Feng,
  Junhong Huang, and Baoxun Wang. 2020.
\newblock Guiding variational response generator to exploit persona.
\newblock In \emph{Proceedings of the 58th Annual Meeting of the Association
  for Computational Linguistics}, pages 53--65, Online. Association for
  Computational Linguistics.

\bibitem[{Zhang et~al.(2021)Zhang, Li, Shen, Xie, and
  Qian}]{Zhang_Li_Shen_Xie_Qian_2021}
Chao Zhang, Zhijian Li, Zebang Shen, Jiahao Xie, and Hui Qian. 2021.
\newblock \href {https://doi.org/10.1609/aaai.v35i12.17295} {A hybrid
  stochastic gradient hamiltonian monte carlo method}.
\newblock \emph{Proceedings of the AAAI Conference on Artificial Intelligence},
  35(12):10842--10850.

\bibitem[{Zhang et~al.(2020)Zhang, Sun, Galley, Chen, Brockett, Gao, Gao, Liu,
  and Dolan}]{zhang-etal-2020-dialogpt}
Yizhe Zhang, Siqi Sun, Michel Galley, Yen-Chun Chen, Chris Brockett, Xiang Gao,
  Jianfeng Gao, Jingjing Liu, and Bill Dolan. 2020.
\newblock \href {https://doi.org/10.18653/v1/2020.acl-demos.30} {{DIALOGPT} :
  Large-scale generative pre-training for conversational response generation}.
\newblock In \emph{Proceedings of the 58th Annual Meeting of the Association
  for Computational Linguistics: System Demonstrations}, pages 270--278,
  Online. Association for Computational Linguistics.

\bibitem[{Zhao et~al.(2017)Zhao, Zhao, and Eskenazi}]{zhao-etal-2017-learning}
Tiancheng Zhao, Ran Zhao, and Maxine Eskenazi. 2017.
\newblock Learning discourse-level diversity for neural dialog models using
  conditional variational autoencoders.
\newblock In \emph{Proceedings of the 55th Annual Meeting of the Association
  for Computational Linguistics (Volume 1: Long Papers)}, pages 654--664,
  Vancouver, Canada. Association for Computational Linguistics.

\bibitem[{Zhou and Wang(2018)}]{zhou-wang-2018-mojitalk}
Xianda Zhou and William~Yang Wang. 2018.
\newblock \href {https://doi.org/10.18653/v1/P18-1104} {{M}oji{T}alk:
  Generating emotional responses at scale}.
\newblock In \emph{ACL}, pages 1128--1137, Melbourne, Australia. Association
  for Computational Linguistics.

\end{thebibliography}
\bibliographystyle{acl_natbib}

\clearpage

\appendix

\section{Appendix}
\label{sec:appendix}

\subsection{Corpora}
For our experiments, we utilise the DailyDialogs and EmpatheticDialogs corpora. A summary of the number of dialogues available for training, validation, and testing are provided in Table 4. Both corpora provide additional labels depicting the emotion, topic etc. However, for our experiments, all additional labels corresponding to each dialogue are not utilized, only the dialogue utterances are used.

\subsection{Evaluation Details}

\subsubsection{Diversity Metrics}
To measure diversity, we utilize the inter-response Distinct-$1$ and $2$ scores \cite{li-etal-2016-diversity}, which accounts for the number of unique $1$ or $2$-grams in the generated response. A higher distinct score indicates greater overall response diversity. We also employ traditional lexical diversity metrics such as the Textual Lexical Diversity (MTLD) \cite{fergadiotis2011modeling}, the Moving-Average Type–Token Ratio (MATTR) \cite{covington2007mattr}, and the Hypergeometric Distribution Diversity (HDD) index \cite{mccarthy2007vocd} from the field of linguistics to measure the  corpus-level diversity. The MATTR score is the average of Token-Type-Ratio (TTR) of each segment of the response with a fixed window size $w=50$. The MTLD score reflects the TTR of sequentially larger segments of the response until a fixed threshold $h=0.72$. The HDD index is the sum of the probabilities of finding each token in a random sample of $n = 42$ words taken from the response.

\subsubsection{Coherence Metrics}

To measure coherence, \citet{9747458} presented the Utterance Entailment (UE) score. Essentially, computing the UE score involves applying a BERT-based Natural Language Inference (NLI) model to the generated response and each utterance in the dialogue context. A score of 1,-1 or 0 is assigned when the response and utterance are either entailing, contradictory or neutral respectively. The UE score is computed by averaging all assigned ratings. However, the length and semantic content of each utterance could affect the quality of the predictions by the NLI model. Extremely long, multi-sentence utterances could result in low accuracy predictions, and cursory utterances such as 'thank you' or 'no problem' would further dilute the final score. Hence, in our implementation of the UE score, each utterance is split into individual sentences. Then, to remove cursory segments, sentences which consist of fewer than four words are removed. The remaining sentences are fed to the NLI model alongside the generated response, and the individual ratings are collated. The UE score is the average of all collated ratings.
\begin{table}[]
\centering
\caption{breakdown of the number of dialogues available in the DailyDialog and EmpatheticDialogs corpora.}
\scalebox{0.7}{
\begin{tabular}{@{}ccc@{}}
\toprule
      & DailyDialog & EmpatheticDialogs \\ \midrule
Train & 11118       & 19533             \\
Valid & 1000        & 2770              \\
Test  & 1000        & 2547              \\ \bottomrule
\end{tabular}}
\end{table}

\begin{table}
\centering
\caption{Automatic evaluation results for the \emph{opposite}, \emph{weights only}, and \emph{bias only} configurations of BODEB\textsubscript{M} (using DialoGPT) on the EmpatheticDialogs corpus.}
\scalebox{0.68}{
\begin{tblr}{
  hline{1-2,7} = {-}{},
}
                       & Dist-1 & Dist-2 & MATTR & MTLD   & HDD   & UE  \\
BODEB\textsubscript{M}              & \textbf{0.056}  & 0.369   & \textbf{0.748} & 48.949 & 0.769 & \textbf{0.245}       \\
-\textit{opposite}     & 0.054  & \textbf{0.372} & 0.741 & \textbf{49.123} & \textbf{0.782} & 0.206*    \\
-\textit{weights only} & 0.050* & 0.271*  & 0.699 & 44.236* & 0.732 & 0.214*     \\
-\textit{bias only}    & 0.053  & 0.338  & 0.738 & 47.816 & 0.771 & 0.202*   \\
-\textit{none}         & 0.051*  & 0.289*  & 0.674* & 36.964* & 0.705* & 0.210*
\end{tblr}}
\end{table}
\begin{table}
\centering
\caption{Automatic evaluation results for DialoGPT with temperature ($T$) adjustment on DailyDialog}
\scalebox{0.68}{
\begin{tblr}{
  hline{1-2,8} = {-}{},
}
                       & Dist-1 & Dist-2 & MATTR   & MTLD   & HDD   & UE \\
BODEB\textsubscript{M}               & 0.056  & 0.369  & \textbf{0.748} & \textbf{48.949} & 0.769 & \textbf{0.245}       \\
DialoGPT                & 0.043*  & 0.207*  & 0.653* & 31.547* & 0.694* & 0.233       \\
DialoGPT\textsubscript{T=0.25}       & 0.047*  & 0.216*  & 0.649* & 33.684* & 0.681* & 0.227*    \\
DialoGPT\textsubscript{T=0.5}          & 0.049*  & 0.222*  & 0.669* & 36.101* & 0.709* & 0.231    \\
DialoGPT\textsubscript{T=0.75}         & 0.054  & 0.306*  & 0.723 & 47.915 & \textbf{0.776} & 0.219*     \\
DialoGPT\textsubscript{T=1.0}          & \textbf{0.060}  & \textbf{0.378}  & 0.731 & 48.239 & 0.761 & 0.186*   \\
\end{tblr}}
\end{table}

\begin{table*}[]
\centering
\caption{Samples of dialogue contexts and responses generated by DialoGPT-based models from the DailyDialog corpus. }
\scalebox{0.8}{
\begin{tabular}{|c|l|l|}
\hline
\multirow{3}{*}{\begin{tabular}[c]{@{}c@{}}Dialogue\\ Context\end{tabular}}    & User:     &Here's a model of our latest design. What do you think of it?  \\
                                                                               & Agent:    &Well, I'm not too happy about it.  \\
                                                                               & User:     &What's wrong with it?  \\ \hline
\multirow{5}{*}{\begin{tabular}[c]{@{}c@{}}Generated\\ Responses\end{tabular}} & DialoGPT: &I don't like it.  \\
                                                                               & CVAE:     &I think it can be a good one.  \\
                                                                               & UA-CVAE:  &It is terrible.  \\
                                                                               & MOPED:  &I am not one or the other person, or both. \\
                                                                               & BODEB\textsubscript{G}:     &Oh, I see that it's a bit out of place in the market. But I don't like it.  \\
                                                                               & BODEB\textsubscript{M}:     &I think the design is too wide and heavy for my liking.  \\ \hline
\begin{tabular}[c]{@{}c@{}}Dialogue\\ Context\end{tabular}                     & User:     &I can't stand the stupid guy any longer. It's unbelievable!  \\ \hline
\multirow{5}{*}{\begin{tabular}[c]{@{}c@{}}Generated\\ Responses\end{tabular}} & DialoGPT: &I know, He's so annoying.  \\
                                                                               & CVAE:     &I'm not a little worried about it.  \\
                                                                               & UA-CVAE:  &She told me that he is not here to help us.  \\
                                                                               & MOPED:  &This is a new new movie.  Excuse me?\\
                                                                               & BODEB\textsubscript{G}:     &What's wrong with him?  \\
                                                                               & BODEB\textsubscript{M}:     &What a surprise! He's so bad!  \\ \hline
\end{tabular}}
\end{table*}

\subsubsection{Human Evaluation}
For human evaluation, we compare responses generated by BODEB against responses generated by the pretrained model, CVAE, and MOPED based on `Diversity', `Fluency', and `Coherence'. We recruited five participants for this evaluation. All participants were native English-speaking university graduates of ages between 24 to 27. The evaluation was conducted in-person. The participants were provided with the dialogue samples and told to evaluate the samples separately i.e., there was no contact between participants. Every participant was assigned 50 randomly selected dialogue examples with a response generated by each of the four baselines. Then, they were told to compare responses (the participants were not aware which model generated each response), and indicate if the response generated by BODEB either wins, losses or ties with the other responses. Each participant took approximately one hour to finish the evaluation.

\subsection{Dialogue Samples}
Samples of responses generated by DialoGPT-based models (DialoGPT, CVAE, UA-CVAE, MOPED, BODEB\textsubscript{G}, and BODEB\textsubscript{M}) from the DailyDialog corpus are provided in Table 8.

\subsection{Additional Configuration}

We also attempted to initialise the variance in the opposite direction (\emph{opposite}). We found that this would result in lower response coherence compared to BODEB\textsubscript{M} despite achieving comparable response diversity. This can be inferred from the higher scores attained on diversity metrics and the lower UE-score which is a measure for coherence. This supports our finding that constraining the variance of deeper parameters would improve coherence. In addition, we constructed an evaluated a variant (\emph{weights only}) of BODEB\textsubscript{M} where only the variance of the weights is set in the manner described in section. Variance of the biases are set as per MOPED. We also constructed a flipped variant (\emph{bias only}) where only the variance of the biases is in accordance with BODEB\textsubscript{M}, and the variance of the weights are set as per MOPED. Finally, we implement Bayesian model (\emph{none}) where the variances of both the weights and biases are set as per MOPED. Upon closer inspection of the generated responses and the automatic evaluation scores attained, we similarly found that both variants demonstrated a noticeable drop in contextual coherence despite achieving comparable results in terms of diversity. This further emphasises the effectiveness of BODEB when it comes to preserving coherence. The results attained by the aforementioned models are presented in Table 6.

\subsection{Comparison with Temperature}
We also adjust the temperature parameter ($T$) of DialoGPT. Selecting a larger temperature value would increase randomness and improve diversity at the expense of coherence and vice versa. Automatic evaluation results when $T=0.25,0.5,0.75,1.0$ are presented in Table 7. As evidenced by the increasing diversity scores, as $T$ increases, response diversity improves. Concurrently, based on the UE scores attained, it is also apparent that response coherence drops as as $T$ increases. On the other hand, BODEB\textsubscript{M} achieved response diversity comparable with $T=0.75$ and $T=1.0$ while maintaining response coherence.

\end{document}